\documentclass{article}
\PassOptionsToPackage{numbers, compress}{natbib}
 \usepackage[dblblindworkshop, final]{neurips_2025}
\workshoptitle{Structured Probabilistic Inference and Generative Modeling}

\usepackage[utf8]{inputenc} 
\usepackage[T1]{fontenc}    
\usepackage{hyperref}       
\usepackage{url}            
\usepackage{booktabs}       
\usepackage{amsfonts}       
\usepackage{nicefrac}       
\usepackage{microtype}      
\usepackage{xcolor}         
\usepackage{amssymb}
\usepackage{amsmath}
\usepackage{amsthm}
\usepackage{algorithm}
\usepackage{algpseudocode}
\usepackage[pdftex]{graphicx}
\usepackage{tikz}
\usepackage{ulem}
\newtheorem{theorem}{Theorem}
\newtheorem{lemma}{Lemma}

\newtheorem*{remark}{Remark}
\newtheorem{definition}{Definition}
\newtheorem{assumption}{Assumption}

\usepackage{listings}
\usepackage{cancel}

\usepackage{xcolor}
\definecolor{codegreen}{rgb}{0,0.6,0}
\definecolor{codegray}{rgb}{0.5,0.5,0.5}
\definecolor{codepurple}{rgb}{0.58,0,0.82}

\lstdefinestyle{mystyle}{
    commentstyle=\color{codegreen},
    keywordstyle=\color{magenta},
    numberstyle=\tiny\color{codegray},
    stringstyle=\color{codepurple},
    basicstyle=\ttfamily\footnotesize,
    breakatwhitespace=false,         
    breaklines=true,                 
    captionpos=b,                    
    keepspaces=true,                 
    numbers=left,                    
    numbersep=5pt,                  
    showspaces=false,                
    showstringspaces=false,
    showtabs=false,                  
    tabsize=2
}

\usetikzlibrary{shapes,decorations,arrows,calc,arrows.meta,fit,positioning}
\tikzset{
    -Latex,auto,node distance =1 cm and 1 cm,semithick,
    state/.style ={ellipse, draw, minimum width = 0.7 cm},
    point/.style = {circle, draw, inner sep=0.04cm,fill,node contents={}},
    bidirected/.style={Latex-Latex,dashed},
    el/.style = {inner sep=2pt, align=left, sloped}
}

\title{Towards Practical Multi-label Causal Discovery in High-Dimensional Event Sequences via One-Shot Graph Aggregation}

\author{
Hugo Math$^{1,2}$ \quad Rainer Lienhart$^{2}$ \\
\\
$^1$ BMW Group \\
$^2$ Chair for Machine Learning and Computer Vision, Augsburg University \\
Augsburg, Germany \\
\texttt{hugo.math@bmwgroup.de, rainer.lienhart@uni-augsburg.de}
}

\begin{document}

\maketitle

\begin{abstract}
Understanding causality in event sequences where outcome labels such as diseases or system failures arise from preceding events like symptoms or error codes is critical—yet remains an unsolved challenge across domains like healthcare or vehicle diagnostics. We introduce CARGO, a scalable multi-label causal discovery method for sparse, high-dimensional event sequences consisting of thousands of unique event types. Using two pretrained causal Transformers as domain-specific foundation models for event sequences—CARGO infers in parallel, per sequence, one-shot causal graphs and aggregates them using an adaptive frequency fusion to reconstruct the global Markov boundaries of labels. This two-stage approach enables efficient probabilistic reasoning at scale while avoiding the intractable cost of full-dataset conditional-independence testing.
Our results on a challenging real-world automotive fault prediction dataset with over 29,100 unique event types and 474 imbalanced labels demonstrate CARGO’s ability to perform structured reasoning.

\end{abstract}

\section{Introduction}
Understanding \textit{why} specific events lead to particular outcomes is vital for effective diagnosis, predictions, and overall decision-making \cite{liu2025learning, shtp}. For instance, ''what series of events captured by diagnostic led to this vehicle failure'' or ''what symptoms led to this disease'' \cite{MANOCCHIO2024122564, MedBERT, bihealth, flight_service_cd, pdm_dtc_feature_extraction}. Here, an event sequence consists of a list of discrete events \(x_i\) recorded asynchronously over time, while labels \(\boldsymbol{y}\) summarize outcomes associated with the full sequence (e.g, a diagnosed defect or condition).

A fundamental obstacle in these settings is dimensionality. Real-world systems often involve tens of thousands of possible events, rendering causal discovery intractable for current algorithms \cite{hasan2023a}.
To address this, we reinterpret multi-label causal discovery for event sequences as a form of Bayesian model averaging \cite{HOET1999, pearl_1998_bn}, where each sequence is treated as a sample from a local causal model. Specifically, each sequence induces a one-shot causal graph (i.e., a directed acyclic graph (DAG)) \cite{tf_causalinterpretation_neurips_2023}. Together, they can be fused to form a unified global structure \cite{divide_conquer_mb_discovery}. This process, known as structural fusion \cite{consensus_bn_jose}, aggregates local graphs into a consensus causal graph over all observed sequences.

To summarize our contributions: we introduce CARGO (\uline{C}ausal \uline{A}ggregation via \uline{R}egressive \uline{G}raph \uline{O}perations), the first method to provide scalable causal discovery across thousands of labelled event sequences, with theoretical guarantees under standard assumptions. It is divided into two phases: (1) One-shot graph extraction, where for each sequence, CARGO infers the local Markov boundary of each label using two autoregressive Transformers as density estimators \cite{tf, density_estimator_2O21_journal_ci} (2) Graph fusion, where the local graphs are aggregated via an adaptive threshold function to provide global Markov Boundaries. 

We empirically validate CARGO on a large-scale vehicular dataset comprising about \(29,100\) events and \(474\) imbalanced labels, demonstrating, for the first time, scalability and practical superiority over traditional causal discovery baselines. We also perform ablation on scoring criteria, frequency thresholds, and Transformer quality.

\section{Preliminary \& Related Work}
A full description of the notations and definitions used throughout the paper can be found in Appendix \ref{appendix:not_def}.

\textbf{Event Sequence Modelling.}
Event sequences are typically represented as a series of time-stamped discrete events \(S = \{(t_1, x_1), \ldots, (t_L, x_L)\}\) where \(0 \leq t_1 < \ldots \leq t_L\) denotes the time of occurrence of event type \(x_i \in \mathbb{X}\) drawn from a finite vocabulary \(\mathbb{X}\). In multi-label settings, a binary label vector \(\boldsymbol{y} \in \{0, 1\}^{|\mathbb{Y}|}\) is attached to \(S\) and indicates the presence of multiple outcome labels chosen from \(\mathbb{Y}\) occurring at the final time step \(t_L\). Together, this results in a multi-labeled sequence \(S_l = (S, (\boldsymbol{y}_L, t_L))\). 

Event sequence modelling has been widely applied to predictive tasks. For instance, in the automotive domain, Diagnostic Trouble Codes (DTCs) \cite{pdm_dtc_feature_extraction} are logged asynchronously over time and used to infer failures or error patterns \cite{math2024harnessingeventsensorydata}. In healthcare, electronic health records encode temporal sequences of symptoms to perform predictive tasks \cite{MedBERT, pmlr-v219-labach23a, bihealth}. A common modelling strategy  \cite{crf, cmm} separates such event types \(\mathbb{X}\) from labels \(\mathbb{Y}\).

Transformers \cite{tf, gpt, touvron2023llamaopenefficientfoundation} have emerged as the dominant architecture for sequence modelling. 
Recent work leveraged Transformers in high-dimensional event spaces for next-event and label prediction. \citet{math2024harnessingeventsensorydata} proposed a dual Transformer architecture in which one model predicts the next event type (DTC) and the other predicts label occurrence (e.g., error patterns). Through this paper, we repurpose this dual architecture for causal discovery. 

\textbf{Multi-label Causal Discovery}
 seeks to identify the Markov Boundary (\textbf{MB}) of each label—its minimal set of parents, children, and spouses—such that the label is conditionally independent of all other variables given its \textbf{MB} \cite{optimal_feature_set_cd} (Def. \ref{def:markov_boundary}). 

While classical constraint-based algorithms have shown success on low-dimensional tabular data \cite{constrainct_based_cd, causality_based_feature_selection_2019}, their application to event sequences with multi-label outputs remains challenging due to: (1) \textit{dimensionality}—thousands of event types increase super-exponentially the number of graphs (2) \textit{sparsity}—multi-hot encodings often underrepresent rare but important events (3) \textit{distributional assumptions}—such as linearity or Gaussian noise, which rarely hold in real-world sequences \cite{cd_temporaldata_review}. 

Contemporary work points to decomposing classical causal discovery for high-dimensional datasets into sub-problems and graph aggregation. \citet{laborda_ring_based_distirbuted} introduce a ring-based distributed algorithm for learning high-dimensional BN, \citet{divide_conquer_mb_discovery} explores a distributed approach for large-scale causal structure learning and \citet{recursive_mb_approach_pmlr_large_scale} for Markov Boundaries.

\textbf{Bayesian Network} \cite{pearl_1998_bn} has served as a modelling technique for a variety of decision problems. It is defined as a triplet \(<\boldsymbol{U}, \mathbb{G}, P>\) with \(P\) the joint distribution over a variable set \(\boldsymbol{U}\) of a directed acyclic graph \(\mathbb{G} = (\boldsymbol{U}, E)\) with \(E\) as the set of directed edges. This triplet must satisfy the Markov Condition: every random variable \(U_i\) is independent of its non-descendant variables given its parents \(\text{Pa}(U_i)\) in \(\mathbb{G}\). The directed edge \((U_i \rightarrow U_j)\) encodes a probabilistic dependence. Thus, the joint probability distribution can be factorized as:

\[P(U_1, \cdots, U_n) = \prod^n_{i=1} P(U_i|\text{Pa}(U_i))\]

The DAG encodes a set of conditional independencies \(\mathcal{I}(\mathbb{G})\), where each element corresponds to a conditional independence relation \(U_i \perp U_j | \boldsymbol{Z}\), meaning that \(U_i\) and \(U_j\) are conditionally independent given the set of variables \(\boldsymbol{Z}\).
Formally, a DAG \(\mathbb{G}_k\) is an I-map (or Independence map) of another \(\mathbb{G}\) if the set of conditional independencies encoded by \(\mathbb{G}_k\) is a subset of those encoded by \(\mathbb{G}\): \[\mathcal{I}(\mathbb{G}_k) \subseteq \mathcal{I}(\mathbb{G})\]

\(\mathbb{G}_k\) is a minimal I-map of \(\mathbb{G}\) if removing any edge from it introduces a conditional dependence that would violate an independence in \(\mathbb{G}\), i.e \(\mathcal{I}(\mathbb{G}_k \textbackslash \{e\} \not\subseteq I(\mathbb{G}) \forall e \in E\).

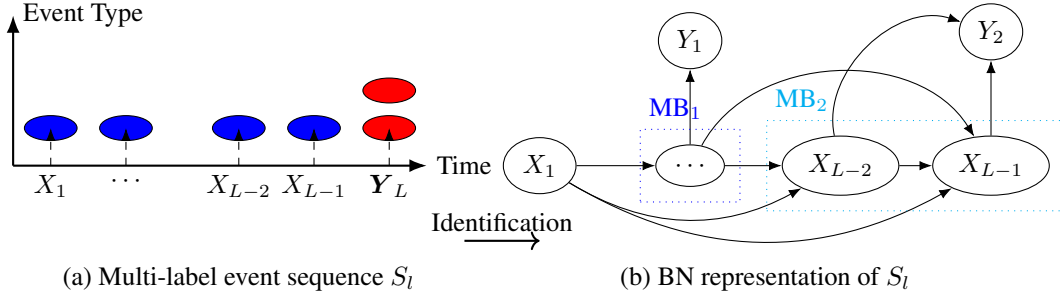
\begin{figure}[!b]
    \centering
    \caption{An example of a causal graph extracted from a single multi-label \textcolor{blue}{event} sequence where \textcolor{blue}{\(\text{MB}_1\)} represents the Markov Boundary of \textcolor{red}{\(Y_1\)} and \textcolor{cyan}{\(\text{MB}_2\)} the Markov Boundary of \textcolor{red}{\(Y_2\)}.}
    \label{fig:markov_boundary_identification}

    \begin{tikzpicture}

        \draw[thick] (-2,0) -- (3.5,0) node[right] {Time};
        \draw[thick] (-2,0) -- (-2,2) node[right] {Event Type};

        \node[state,fill=blue] (x1) at (-1.5,0.5) {};
        \node[state,fill=blue] (x2) at (-0.5,0.5) {};
        \node[state,fill=blue] (x3) at (1,0.5) {};
        \node[state,fill=blue] (x4) at (2 ,0.5) {}; 

        \node[state,fill=red] (y) at (3,0.5) {}; 
        \node[state,fill=red] (y) at (3, 1) {}; 

        \draw[dashed] (-1.5,0) -- (-1.5,0.5);
        \draw[dashed] (-0.5,0) -- (-0.5,0.5);
        \draw[dashed] (1,0) -- (1,0.5);
        \draw[dashed] (2,0) -- (2,0.5);
        \draw[dashed] (3,0) -- (3,0.5);

        \node[below] at (-1.5,0) {\(X_1\)};
        \node[below] at (-0.5,0) {\(\cdots\)};
        \node[below] at (1,0) {\(X_{L-2}\)};
        \node[below] at (2,0) {\(X_{L-1}\)};
        \node[below] at (3,0) {\(\boldsymbol{Y}_{L}\)};

        \draw[thick,->] (4,-1) -- (5,-1) node[midway,above] {Identification};


        \node[state] (cx1) at (5,0) {\(X_{1}\)};
        \node[state] (cxd) at (7,0) {\(\cdots\)};
        \node[state] (cx2) at (9,0) {\(X_{L-2}\)};
        \node[state] (cxi) at (11,0) {\(X_{L-1}\)};
        \node[state] (cy) [above =of cxd] {\(Y_1\)};
        \node[state] (cy2) [above =of cxi] {\(Y_2\)};

        \path (cx1) edge (cxd);
        \path (cxd) edge  (cy);
        \path (cx1) edge[bend right=30] (cx2);
        \path (cx1) edge[bend right=30] (cxi);
        \path (cx2) edge (cxi);
        \path (cx2) edge[bend left=60] (cy2);
        \path (cxd) edge (cx2);
        \path (cxi) edge (cy2);
        \path (cxd) edge[bend left=60] (cxi);

        \node[draw=blue,dotted,fit=(cxd) (cxd), inner sep=0.2cm] (mb1) {};
        \node[anchor=south west, blue] at (mb1.north west) {\(\text{MB}_1\)};
        \node[draw=cyan,dotted,fit=(cx2) (cxi), inner sep=0.2cm] (mb2) {};
        \node[anchor=south west, cyan] at (mb2.north west) {\(\text{MB}_2\)};

        \node at (1,-1.5) {(a) Multi-label event sequence \(S_l\)};
        \node at (8,-1.5) {(b) BN representation of \(S_l\)};

    \end{tikzpicture}
\end{figure}

\textbf{Bayesian Model Averaging.}
Fusing BN has several direct applications. Either to average multiple models from different experts to learn a global and average representation \cite{HOET1999}. Or to perform causal discovery in distributed settings with federated learning algorithms \cite{fedpc_fedfci, Guo_Yu_Liu_Li_2024}.

Formally, 
Given a set of Bayesian Networks \(\{B_k\}^m_{k=1}\) with associated DAGs \(\{\mathbb{G}_k = (V_k, E_k)\}^m_{k=1}, V_k \in \boldsymbol{U}\) sharing the same finite set of node \(\boldsymbol{U}\), structural fusion aim to construct the DAG \(\mathbb{G}^* = (V, E), V \in \boldsymbol{U}\). Multiple  fusion methods exist and leverage either the probability distribution \(p\) by doing Bayesian Model Averaging \cite{HOET1999} or focus on the structural learning (Fig. \ref{fig:structural_fusion_example}) of \(\mathbb{G}^*\) \cite{delSagrado2001, consensus_bn_jose, go2022robust, approx_fusionPUERTA2021155, Guo_Yu_Liu_Li_2024}, which is an NP-hard \cite{consensus_bn_jose} problem.

We will focus on the second element in this paper.
Hence, we are seeking the merged edges \(E = \bigcup^m_{i=1} E^\sigma_i\). The consistent node ordering \(\sigma\) ensures acyclicity. The fused DAG \(\mathbb{G}^*\) is the minimal I-map of the intersection of the conditional independencies across all DAGs \(\mathbb{G}_k = (V_k, E_k)\}^m_{k=1}\).

\textbf{Greedy Equivalence Search.}
(GES) \cite{ges} is one of the most theoretically sound methods to recover a Markov equivalence class (MEC, Def. \ref{def:mec}) of a DAG. In large-sample settings, GES provides theoretical guarantees of recovering the true graph.
Formally, GES searches for the MEC of the graph \(\mathbb{G}^*\) from the observational dataset \(\boldsymbol{D} \) with distribution \(p\). This defines the optimization problem as:

\begin{equation}\label{eq:ges_opti}
    \mathbb{G}^* = \arg_{\mathbb{G}} \text{max} \; S(\mathbb{G}, \boldsymbol{D})
\end{equation}
\citet{ges} proved that under parametric assumption, a large number of samples and using the Bayesian Information Criterion (BIC) as criterion \(S\), GES is guaranteed to recover a Markov Equivalence Class of \(\mathbb{G}^*\).

 

\textbf{Frequency Based.}
Multiple heuristics have been developed to merge multiple BNs. 
One is edge frequency cutoff \cite{Steele2008Consensus}, 
other on estimating the proportion of false positive edges  \cite{consensus_bn_freq_cutoff_no} and integer linear programming (ILP) \cite{consensus_bn_freq_cutoff_no} or a mix of both: \cite{torrijos2025informedgreedyalgorithmscalable}. As pointed out by \cite{consensus_bn_freq_cutoff_no}, choosing a frequency cutoff \(\tau\) is particularly challenging. Moreover, under class imbalance and long-tail problem in classification \cite{longtailissue}, the theoretical property of frequency approaches, such as the law of large numbers, might not hold.

We will now explain the two phases of CARGO, which are (1) One-shot causal discovery (2) Graph aggregation using adaptive thresholds. The Proofs of Lemmas and Theorems can be found in the Appendix \ref{appendix:proofs}.

\section{One-Shot Causal Discovery}\label{sec:oneshotmarkov}
Let \(S^k_l\) be a multi-labeled sequence drawn from a dataset \(D = \{S^1_l, \cdots, S^m_l\} \subset \mathbb{S}\) and  \(\mathbb{G}_k\) the sequential BN (Fig. \ref{fig:markov_boundary_identification}) with attached labels. The goal of multi-label causal discovery is to identify the Markov Boundary of each label \(Y_j \in \boldsymbol{Y}\) present in \(S^k_l\). 
To be able to access conditional independence (Def. \ref{def:conditional_independence}) between \(X_i\) and \(Y_j\) conditioned on the past events \(\boldsymbol{Z} = (x_1, \cdots, x_{i-1}) = S_{< i}\), we model the event apparitions using a sequential BN (Fig. \ref{fig:markov_boundary_identification}). 

We want to assess how much additional information event \(X_i\) occurring at step \(i\) provides about label \(Y_j\) when we already know the past sequence of events \(\boldsymbol{Z} = S_{<i}\). We essentially try to answer if:
\[P(Y_{j}|X_i, \boldsymbol{Z}) = P(Y_{j}|\boldsymbol{Z}) \Leftrightarrow 
D_{KL}(P(Y_{j}|X_i, \boldsymbol{Z})\|P(Y_{j}|\boldsymbol{Z})) = 0\]
where \(D_{KL}\) denotes the \textit{Kullback-Leibler divergence} \cite{cover1999elements}. The distributional difference between the conditionals \(P(Y_j|X_i, \boldsymbol{Z}), P(Y_j| \boldsymbol{Z})\) is akin to Information Gain \(I_G\) \cite{quinlan:induction} conditioned on past events:

\begin{equation}\label{eq:info_gain}
    I_G(Y_j, x_i|z) \triangleq D_{KL}(P(Y_j|X_i=x_i, \boldsymbol{Z} = z)) || P(Y_j|\boldsymbol{Z}=z)) 
\end{equation}
Which is equals to the difference between the conditional entropies \cite{cover1999elements} denoted as \(H\): 
\begin{equation}
    I_G(Y_j, x_i|z) = H(Y_j|z) - H(Y_j|x_i, z)
\end{equation}

More generally, we can access the conditional independence of event \(X_i\) and label \(Y_{j}\) using the conditional mutual information (CMI) \cite{cover1999elements} which is simply the expected value over \( z\) of the information gain \(I_G(Y_j, X_i|z)\) such as:
\begin{equation}
    I(Y_{j}, X_i|\boldsymbol{Z}) \triangleq H(Y_j|\boldsymbol{Z}) - H(Y_j|\boldsymbol{Z}, X_i)\label{eq:cmi_theorique}
    = \mathbb{E}_{p(z)}[I_{G}(Y_j, X_i|\boldsymbol{Z}=z)]
\end{equation}

It can be interpreted as the expected value over all possible contexts \(\boldsymbol{Z}\) of the deviation from independence of \(X_i, Y_j\) in this context. To approximate  Eq.~\eqref{eq:cmi_theorique},
 a naive Monte Carlo \cite{Doucet2001} approximation is performed where we draw $N$ random variations of the conditioning set 
\(z^{(l)} = \{x^{(l)}_0, \ldots, x^{(l)}_{i-1}\}\), denoting the $l$-th sampled \emph{particle}:
\begin{align}\label{eq:cmi_approx}
\hat{I}_N(Y_{j}, X_i \mid \boldsymbol{Z}) 
&= \frac{1}{N} \sum_{l=1}^N I_G(Y_{j}, X_i \mid \boldsymbol{Z} = z^{(l)})
\end{align}
This estimator is unbiased because the contexts \(z^{(l)}\) are sampled directly from \(\text{Tf}_x\) using a proposal \(Q\) with the same support as \(P(\boldsymbol{Z})\).
Since \(I_G(Y_{j}, X_i \mid \boldsymbol{Z} = z)\) is a difference between conditional entropies (\eqref{eq:info_gain}), it is thus bounded uniformly \cite{cover1999elements} by the log of supports such as:

\[0 < I_G(Y_{j}, X_i \mid \boldsymbol{Z} = z^{(l)}) = H(Y_{j}|z^{(l)}) - H(Y_{j}|X_i, z^{(l)})) \leq H(Y_{j}) \leq \log{|\mathbb{Y}|}\]

Thus the posterior variance of \(f_i =I_G(Y_{j}, X_i \mid \boldsymbol{Z} = z^{(l)})\) satisfies \(\sigma^2_{f_i} \triangleq \mathbb{E}_{p(z)}[f^2_i(p(z)] - I^2(f_i) < +\infty\) \cite{Doucet2001} then the variance of \(\hat{I}_N(f_i)\)) is equal to \(\textit{var}(\hat{I}_N(f_i)) = \frac{\sigma^2_{f_t}}{N}\) and from the strong law of large numbers: 
\begin{align}
\hat{I}_N &\xrightarrow[N \to +\infty]{\text{a.s.}} 
\mathbb{E}_{p(z)}\!\left[ I_G(Y_{j}, X_i \mid \boldsymbol{Z}=z) \right] 
\triangleq I(f_i).
\end{align}

\textbf{Density Estimation.}
We used two pretrained Transformers \((\text{Tf}_x, \text{Tf}_y)\) trained via maximum likelihood on a dataset of multi-labelled event sequences \(D = \{S^1_l, \cdots, S^m_l\} \subset \mathbb{S}\). We assume that they perfectly model the true conditional distributions of events and labels (A\ref{assumption:oracle}). While, due to the strict equivalences between CI-tests and conditional independence, it is difficult to provide a theoretical guarantee under imperfect models, we acknowledge that this assumption may be violated. We note that most causal discovery methods either impose strong parametric data assumptions or rely on a perfect CI-test. We provide an ablation study on the impact of the NADE's quality on the one-shot phase in Appendix \ref{appendix:nades}, including the number of parameters, context \(c\), and \(\text{Tf}_y\) performance.

Formally, the two Transformers infer the probability of the next event and label conditioned on the past events using the hidden states \(\boldsymbol{h}^{x}_{i-1}, \boldsymbol{h}^{y}_{i} \in \mathbb{R}^d\), from \(\text{Tf}_x, \text{Tf}_y\) respectively:
\begin{align}
\text{Tf}_x(S_{< i}) &= \textit{Softmax}(\boldsymbol{h}^{x}_{i-1}) = P_{\theta_x}(X_i| \boldsymbol{Z}) \label{eq:prob_x} \\
\text{Tf}_y(S_{\leq i}) &=  \textit{Sigmoid}(\boldsymbol{h}^{y}_{i}) = P_{\theta_y}(Y|X_i, \boldsymbol{Z})\label{eq:prob_y}
\end{align}

\textbf{Sequential One-shot Causal Discovery.}
The CMI using Eq.\eqref{eq:cmi_theorique} is computable only with the posteriors \(P(Y_j|\boldsymbol{Z}), P(Y_j|X_i, \boldsymbol{Z}))\).
In practice a label-specific threshold \(\theta_j \approx 0\) is applied to Eq.~\eqref{eq:cmi_theorique} to identify conditional dependence:
\begin{equation}\label{eq:cmi_epsilon}
    Y_j \not\!\perp X_i \mid \boldsymbol{Z} \quad \Leftrightarrow \quad I(Y_j, X_i \mid \boldsymbol{Z}) > \theta_j \approx 0.
\end{equation}

Hence, the expectation in Eq\(. \eqref{eq:cmi_theorique}\) is computed using a Monte-Carlo simulation, by sampling \(N\) similar context \(\boldsymbol{Z}\) from \(\text{Tf}_x\). 
Such that for each position in the sequence, we generate \(N\) plausible next tokens using a combination of top-k and nucleus sampling \cite{Holtzman2020The}. Ablation studies on the effect of the sampling method and thresholds are given in Appendix \ref{sec:samplingnumber}, \ref{sec:dynamic_threshold}.

\begin{theorem}[Markov Boundary Identification in Event Sequences]
\label{th:mb-recovery}
If  \(S^k_l\) a multi-labeled sequence drawn from a dataset  \(D = \{S^1_l, \cdots, S^m_l\} \subset \mathbb{S}\) where two Oracle Models \(\text{Tf}_x\) and \(\text{Tf}_y\) were trained on, then under causal sufficiency (A\ref{assumption:causal_sufficiency}), bounded lagged effects (A\ref{assumption:lagged_effects}) and temporal precedence (A\ref{assumption:temporal_precedence}), the Markov Boundary of each label \(Y_j\) in the causal graph \(\mathbb{G}\) can be identified using conditional mutual information for CI-testing. 
\end{theorem}

Theorem \ref{th:mb-recovery} enables us to sequentially recover the Markov Boundary of each label in a sequence. It provides a theoretical guarantee to recover the correct causes for each label \(Y_j\).

\begin{figure}[!t]
    \centering
    \caption{The overview of CARGO. Phase 1 (One-shot) is on top, and Phase 2 (Adaptive Thresholding) is on the bottom. \(d\) denotes the hidden dimension, \(L\) the sequence length, \(m\) the number of samples and \(\text{MB}_1, \text{MB}_2\) the Markov Boundary of \(Y_1, Y_2\). All green and blue areas are parallelized.}
    \includegraphics[width=1\linewidth]{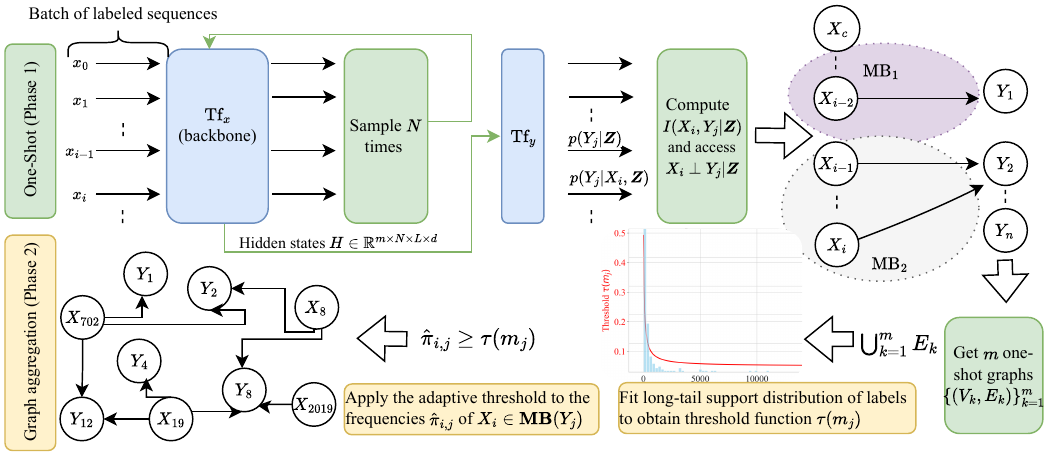}
    \label{fig:cargo}
\end{figure}

\textbf{Computation.}
A key advantage of our approach is its scalability. Unlike traditional methods whose complexity depends on the event and label cardinality \(|\mathbb{X}|\) and \(|\mathbb{Y}|\) \cite{feature_selection_review}, phase 1 is agnostic to both. 
Figure~\ref{fig:cargo} shows all parallelized steps on GPUs. CMI estimations are independently performed for all positions \(i \in [c, L]\), with the sampling pushed into the batch dimension and results averaged across labels. This transitions the time complexity from \(\mathcal{O}(\text{BS} \times N \times L)\) to \(\mathcal{O}(1)\) per batch, with \(L\) being the sequence length.

To ensure stable conditional entropy estimates and reliable predictions from \(\text{Tf}_y\), the CMI is computed after observing \(c\) events (\textit{context}). This design choice also enables out-of-the-box parallelization. By sampling \(N\) variations of the prefix sequence \(S_{\leq c}\), the CMI is independently computed across positions \(i \in [c, L]\). In our experiments, we set \(c=15, L=192\).
The implementation of Phase 1 in \textit{PyTorch} \cite{pytorch} is provided in Appendix~\ref{lst:phase1}.

\section{Structural Fusion of Markov Boundaries}

Let \(\{\mathbb{G}_k = (V_k, E_k)\}^m_{k=1}, V_i \in \boldsymbol{U}\) be the set of DAGs generated by the Phase 1 from the dataset \(\boldsymbol{D}\) containing \(m\) i.i.d sequences \(\{S^k_l\}^m_{k=1}\) drawn from a joint distribution \(p(x,y)\). Each graph \(\mathbb{G}_k\) represents local Markov Boundaries identified within sequence \(S^k_l\). Our objective is to fuse these local graphs into a single, global consensus graph \(\mathbb{G}^* =  (\boldsymbol{U}, E)\) (see Fig. \ref{fig:structural_fusion_example}) with the events always as parents of labels \(Y_j\) such as:
\[\text{Pa}(Y_j) \subseteq \{X_1, \cdots, X_n\}\] 

A naive fusion approach, such as taking the simple union of all edges \(E = \bigcup^m_{k=1} E^\sigma_k\) works iff the Oracle models yield the perfect CI-tests in Phase 1 and thus the local graphs \(\mathbb{G}^k\) are faithful to \(p(x,y)\)  (\cite{consensus_bn_jose}, Theorem 4.). 
Here, the ordering \(\sigma\) doesn't matter since we are dealing with Markov Boundaries. Moreover, \(G^*\) is naturally a DAG because we considered previously that outcome labels are solely explained by events, which simplifies acyclicity. We define the Bernoulli variable \(Z^{k}_{i,j}\) for each potential edge \((X_i \rightarrow Y_j)\) within each sequence \(S^k_l\):
\[Z^k_{i,j} = \begin{cases}
    1 \; \text{if the edge} \; X_i \rightarrow Y_j \; \text{is present in } \; \mathbb{G}_k \\
    0 \; \text{otherwise}
\end{cases}\]
Under the Oracle Models assumption (A\ref{assumption:oracle}), our one-shot discovery phase serves as a perfect conditional-independence tester. Consequently, the detection of an edge in a local graph \(\mathbb{G}_k\) corresponds precisely to a true causal dependency in the global graph \(\mathbb{G}^*\). The probability of this event \(P(Z^k_{i,j} = 1)\) is therefore the true marginal probability of the edge's existence, which we denote as \(\pi_{i,j}\).

The empirical frequency, \(\hat{\pi}_{i,j}(m)\), of the edge \((X_i \rightarrow Y_j)\) after observing \(m\) sequences is the sample mean of these i.i.d Bernoulli variables \(
\hat{\pi}_{i,j}(m) = \frac{1}{m}\sum_{k=1}^{m} Z_{ij}^{k}\).
By the Law of Large Numbers (LLN), as the number of i.i.d sequences \(m\) tends to infinity, the empirical frequency converges in probability to the true expected value of the random variable: 
\[
\hat{\pi}_{i,j}(m) \xrightarrow{p} \mathbb{E}[Z_{i,j}^{(k)}] = \pi_{i,j}
\]
Thus, given a sufficiently large number of sequences, the empirical frequency \(\hat{\pi}_{i,j}(m)\) serves as a consistent estimator for the true probability of the edge's existence in the global DAG \(\mathbb{G}^*\).

\subsection{Aggregation under imperfect CI-tests}\label{sec:agg_imperfect_ci_tests}
The assumption of an Oracle CI-tester, while necessary for initial theoretical guarantees, is invariably violated in practice due to factors like model capacity, limited data, or class imbalance. The extracted one-shot graphs will most likely violate the independencies in \(\mathbb{G}_k\) and thus \(\mathbb{G}^*\).

Let us model the performance of our one-shot CI-test for any potential edge \(X_i \rightarrow Y_j\) with the following error rates: (1) False Positive Rate (Type I Error): \(\alpha = P(\text{detect | edge is spurious)}\) (2) True Positive Rate (Sensitivity): \(1-\beta = P(\text{detect | edge is causal)}\).
We operate under the reasonable assumption that our one-shot classifier is significantly better than random, which implies that \(1-\beta \gg \alpha\).
The expected value of our Bernoulli variable \(Z_{i,j}^k\) is now:
\begin{align*}
\mathbb{E}[Z_{ij}^{k}] &= P(Z_{ij}^{k}=1) \\
&= P(\text{detect}\;|\;\text{causal})P(\text{causal}) + P(\text{detect}\;|\;\text{spurious})P(\text{spurious}) \\
&= (1-\beta)\pi_{ij} + \alpha(1-\pi_{ij})
\end{align*}
The empirical frequency now converges to this new expectation. For a true edge \((\pi_{i,j} = 1)\), the empirical frequency converges to a high value: 
\(
\hat{\pi}_{ij}(m) \xrightarrow{p} 1 - \beta
\) and for a spurious edge \((\pi_{i,j} = 0)\) it converges to a low value: \(\hat{\pi}_{ij}(m) \xrightarrow{p} \alpha\)
This reveals the critical role of frequency aggregation as a mechanism for separating signal from noise.

\subsection{Adaptive Fusion for Structural Discovery in Long-Tail Distributions}
A primary challenge in real-world causal discovery is the long-tail distribution of outcome labels, where a few “head” labels possess abundant data while the vast majority of “tail” labels are data-sparse  \cite{longtailissue}. 

For rare labels, where empirical edge frequencies are high-variance estimators, a conservative high threshold is necessary to maintain precision against statistical noise. Conversely, for common labels where frequencies are reliable, a high threshold would be overly stringent, purging weaker but valid causal links. 
To resolve this, we introduce an \emph{adaptive thresholding} strategy (Fig. \ref{fig:adaptive_threshold}) that tailors the edge inclusion criterion to the statistical power available for each label. We define a label-specific threshold \(\tau_j,\) as a logistic decay function of its sample support \(m_j\): 

\begin{equation}
\tau_j(m_j)= (\tau_{\max} - \tau_{\min}) \cdot \frac{1}{1 + e^{k(\log m_j - \log m_0)}} + \tau_{\min}
\label{eq:adaptive_threshold}
\end{equation}
This function smoothly interpolates between a user-defined maximum threshold, \(\tau_{max}\) (prioritizing precision for the tail), and a minimum, \(\tau_{min}\) (prioritizing recall for the head). Crucially, the function's behavior is calibrated by the distribution of the data. Such that decay midpoint \(m_0\) is set to the median of all label supports, providing a robust anchor point against skew.

The decay rate, \(k\) is made inversely proportional to the log-inter-quartile range of supports such that \(k= \frac{2 \log 3}{\log q_{75} - \log q_{25}}\). 
For rare labels with small \(m_j\), the high variance of the frequency estimate necessitates a high threshold \(\tau_j(m_j)\) that acts as a strong regularizer. For common labels with large \(m_j\), the LLN guarantees the convergence of \(\hat{\pi}_{i,j} \) to the true edge probability, justifying a lower threshold to capture a more complete causal structure. 
Hence, this strategy serves as a data-driven denoising mechanism and shares theoretical parallels with ensembling methods \cite{beiman_decision_trees, consistency_rate_of_dc_tree}, thereby enhancing the robustness and accuracy of the final fused graph.
\begin{figure}
 \caption{Example of an error pattern \((y_1)\) defined as a boolean rules of diagnosis trouble codes \((x_i)\)}\label{fig:ep_def}
\begin{equation}\notag
  y_1 = x_1 \;  \& \; x_5 \;\&\; x_8 \; \& \; x_{18}\; \& \; x_{12} \; \& \; x_3 \; \& \; !x_{10} \;\& \; !x_{20}
\end{equation}
\end{figure}
\section{Empirical Evaluation}
\textbf{Settings \& Vehicle Dataset.}\label{sec:settings}
We used a \(g4dn.12xlarge\) instance from AWS Sagemaker to run comparisons. It contains 48 vCPUs and 4 NVIDIA T4 GPUs. We used a combination of F1-Score, Precision, and Recall with different averaging methods \cite{reviewmultilabellearning} to compare results.
We evaluated our method on a real-world vehicular test set of \(m=300,000\) sequences, with \(|\mathbb{Y}|=474\) different error patterns and \(|\mathbb{X}| = 29,100\) different DTCs forming sequences of \( \approx 100 \pm 35\) events. We used \(\text{Tf}_x\) and \(\text{Tf}_y\) with 90m and 15m parameters \cite{math2024harnessingeventsensorydata}. The two NADEs didn't see the test set during training. 
Domain experts manually define error patterns as Boolean rules among DTCs (Fig. \ref{fig:ep_def}). We set the elements of this rule as the correct Markov Boundary for each label \(y_j\) in the tested sequences. 

\textbf{Multi-label Causal Discovery Comparisons.}
We benchmark CARGO against local structure learning (LSL) algorithms that estimate Markov Boundaries. This includes established approaches such as CMB \cite{cmb}, MB-by-MB \cite{WANG2014252}, PCD-by-PCD \cite{pcdpcd}, IAMB \cite{iamb} from the \textit{PyCausalFS} package \cite{causality_based_feature_selection_2019}, as well as the more recent, state-of-the-art MI-MCF \cite{mimcf}. 
6 random folds of the test data were created and converted into a multi-one-hot data-frame where one row represents one sequence, and each column represents an event type or label \((\mathbb{X, Y})\).

\textbf{Ablation on Aggregation Criterions for Phase 2.}
We provide an Ablation of the different criteria used in the structural fusion of Markov Boundaries.
\emph{Union} stands for a simple union over all edges without any removal.
\emph{Frequency} or edge voting, counts how often is \(X_i \in \text{MB}(Y_j) \). Then apply a static frequency threshold \(\tau = [0.05, 0.25, 0.5, 0.8]\). \emph{MI} uses the mutual information between events and labels as a criterion in a Background Equivalence Search \cite{ges}. \emph{Expected FPR} (false positive ratio) \cite{consensus_bn_freq_cutoff_no} describes two beta distributions which are fitted using the distribution of the mutual information \(I(X_i, Y_j)\) extracted from Phase 1. The lower tail is used for outlier detection. Different FPR  are chosen \(\beta = [0.01, 0.05, 0.15, 0.2]\). 
Detailed definitions can be found in Appendix \ref{appendix:criterions}. 

\subsection{Results}
\textbf{Comparisons.}
We performed comparisons on Table~\ref{tab:performance_comparison} with \(n = 50,000\) random sequences. We found that even under this reduced setup, LSL algorithms failed to compute the Markov Boundaries within 3 days (a 3-day timeout), far exceeding practical deployment limits.
This behavior highlights the current infeasibility of multi-label causal discovery in high-dimensional event sequences.
Current algorithms are cursed under high-dimensional data, since they rely on expensive CI testing that scales quadratically with the number of nodes \cite{cd_temporaldata_review}. This positions CARGO as a more feasible approach for large-scale, multi-label causal discovery.
\begin{table}[h]
\centering
\caption{Comparisons of \textbf{MB} retrieval with \(m=50,000\) samples averaged over \(6-\)folds with \(|Y| = 474, |X|=29,100\) nodes. Averaging is 'weighted'. The symbol ’-’ indicates that the algorithm didn't output the \(\textbf{MBs}\) within 3 days. Metrics are given in \(\%\).}
\begin{tabular}{lcccc}
\hline
\textbf{Algorithm} & \textbf{Precision}↑ & \textbf{Recall}↑ & \textbf{F1}↑ & \textbf{Running Time (min)}↓ \\ \hline
IAMB & - & - & - & \(>4320\) \\
CMB & - & - & - & \(>4320\) \\
MB-by-MB & - & - & - & \(>4320\) \\
PCDbyPCD & - & - & - & \(>4320\) \\
MI-MCF & - & - & - & \(>4320\) \\
CARGO & \(\mathbf{ 60.6 \pm 1.5 }\) & \(\mathbf{ 45.8\pm 1.7}\) & \(\mathbf{ 45.8\pm 1.2}\) & \(\mathbf{11.7}\) \\ 
\hline
\end{tabular}
\label{tab:performance_comparison}
\end{table}

\begin{figure}[!h]
    \centering
    \caption{Comparison of different criteria for the structural fusion (Phase 2) in function of the number of samples \(m\). With \(|Y| = 474, |X|=29,100\) nodes.}\label{fig:ablation_comparaison_criterions}
    \includegraphics[width=0.95\linewidth]{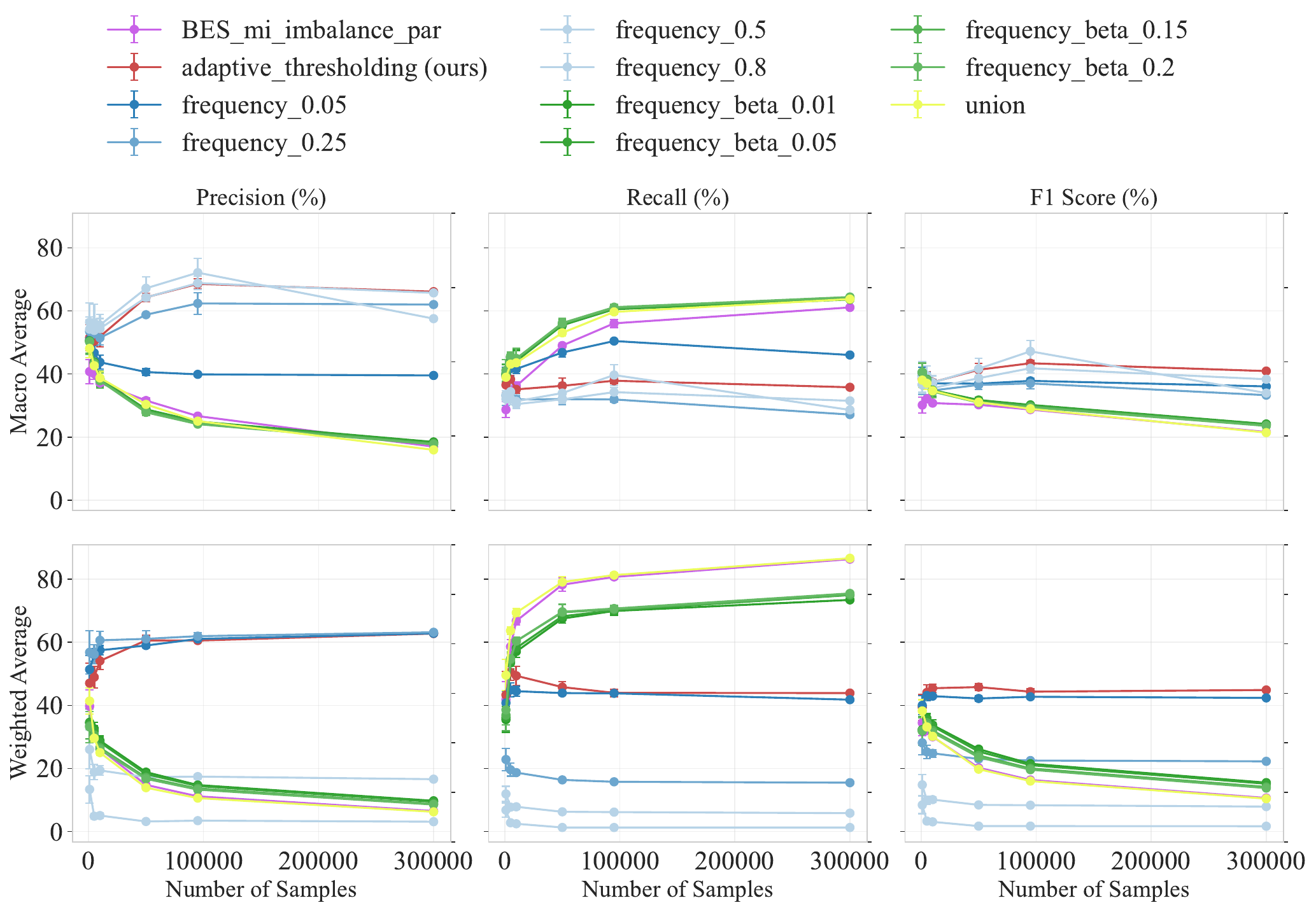}
    \label{fig:placeholder}
\end{figure}

\textbf{Criterions.}
Figure~\ref{fig:ablation_comparaison_criterions} illustrates the impact of aggregation choices during Phase 2. A naïve \emph{Union} maximizes recall (\(84 \%  \; \text{for weighted}\)) but suffers from poor precision. When optimizing a local scoring criterion based on the mutual information \emph{BES mi}, it didn't significantly improve performance over a basic \emph{Union}. 

Moreover, instead of optimizing a score, fitting Beta distributions to detect outliers using their mutual information appears to perform better; hence, \emph{frequency beta} outperforms alternatives, particularly in terms of lower FPR. 
\emph{Frequency} approaches with a static threshold confirms the analysis in Section \ref{sec:agg_imperfect_ci_tests}. When a large number of samples per class \(m_j\) is available, the frequency cut-off \(\tau\) needs to be lower to not penalize classes with big support. Thus, we see that \emph{frequency} with \(\tau=[0.5, 0.8]\) have the lower weighted f1 score of all criterions. On the other hand, a small cut-off \(\tau=[0.05, 0.25]\) enables a huge improvement in the weighted average \((+40\% \; \text{precision})\), but decrease its macro average metrics \((-20\% \; \text{in precision})\). 

Finally, our \emph{adaptive thresholding} criterion leverages a small threshold for big supports and a big threshold for small supports, which takes advantage of the long-tail distribution. As a result, it is first on both averaging, with respectively \(44.88\% \; \text{and} \; 40.9\% \) for weighted and macro f1 score, and \(62.8\% \; \text{and}\: 66.1\%\) for weighted and macro precision. 
\section{Conclusion}
We introduced CARGO, a novel framework for multi-label causal discovery in high-dimensional event sequences. By combining one-shot causal discovery with adaptive frequency-aware aggregation, CARGO successfully recovers interpretable causal structures from noisy observational data—achieving results in minutes where classical methods fail to scale.

CARGO could scale further by leveraging general-purpose foundation models for sequences (e.g., time-series transformers pretrained across domains). Such models could extend the applicability beyond automotive diagnostics to healthcare, cybersecurity, and other structured domains. 

Under the temporal assumptions, large samples, faithfulness, and perfect CI-tests, CARGO recovers the true set of Markov Boundaries. However, we underlined the practical limitations, in particular, long-tail distributions, a common trait in high-dimensional labeled data. 
Future work should extend this framework to support event-to-event causality and, if possible, relax assumptions such as bounded lagged effects or temporal precedence.

Ultimately, CARGO demonstrates how structured probabilistic methods can bridge the gap between causal discovery theory and scalable, practical deployment in complex industrial systems.

\bibliography{math}

\appendix

\newpage
\section{Notations and Definitions}\label{appendix:not_def}
\subsection{Notations}
We use capital letters (e.g., \(X\)) to denote random variables, lower-case letters (e.g., \(x\)) for their realisations, and bold capital letters (e.g., \(\boldsymbol{X}\)) for sets of variables. Let \(\boldsymbol{U}\) denote the set of all (discrete) random variables. We define the event set \(\boldsymbol{X} = \{X_1, \ldots, X_n\} \subset \boldsymbol{U}\), and the label set \(\boldsymbol{Y} = \{Y_1, \ldots, Y_n\} \subset \boldsymbol{U}\). When explicitly said, event \(X^{(t_i)}_i\) represent the occurrence of \(X_i \) at the sequence step \(i\) and time \(t_i\). Similarly for \(Y^{(t_{i+1})}_{i+1}\).

\subsection{Definitions}
\begin{definition}[Faithfulness]\label{def:bn_faithfulness}\citet{Spirtes2001CausationPA}. Given a BN \(<\boldsymbol{U}, \mathbb{G}, P>, \mathbb{G}\) is faithful to \(P\) if and only if every conditional independence present in \(P\) is entailed by \(\mathbb{G}\) and the Markov condition holds. \(P\) is faithful if and only if there exist a DAG \(\mathbb{G}\) such that \(\mathbb{G}\) is faithful to \(P\).
\end{definition}

\begin{definition}[Conditional Independence]\label{def:conditional_independence} Variables \(X\) and \(Y\) are said to be conditionally independent given a variable set \(\boldsymbol{Z}\), if \(P(X, Y|\boldsymbol{Z}) = P(X|\boldsymbol{Z})P(Y|\boldsymbol{Z})\), denoted as \(X \bot \space \space Y| \boldsymbol{Z}\). Inversely, \(X \not\perp \space \space Y| \boldsymbol{Z}\) denotes the conditional dependence.
Using the conditional mutual information \cite{cover1999elements} to measure the independence relationship, this implies that \(\text{I}(X,Y|\boldsymbol{Z}) = 0 \Leftrightarrow X \perp Y | \boldsymbol{Z}\).
\end{definition}
\begin{definition}[Markov Boundary]\label{def:markov_boundary} \citet{optimal_feature_set_cd}.
In a faithful BN \(<\boldsymbol{U}, \mathbb{G}, P>\), for a set of variables \(\boldsymbol{Z} \subset \boldsymbol{U}\) and label \(Y \in \boldsymbol{U}\), if all other variables \(X \in \{\boldsymbol{X} - \boldsymbol{Z}\}\) are independent of \(Y\) conditioned on \(\boldsymbol{Z}\), and any proper subset of \(\boldsymbol{Z}\) do not satisfy the condition, then \(\boldsymbol{Z}\) is the Markov Boundary of \(Y\): \(\textbf{MB}(Y)\).
\end{definition}

\begin{definition}\label{def:mec}
    (Markov Equivalence Class). Two distinct graphs \(\mathbb{G}, \mathbb{G'}\) are said to belong to the same Markov Equivalence Class (MEC) if they have the same set of conditional independencies, i.e \(I(\mathbb{G}) = I(\mathbb{G'})\).
\end{definition}

\begin{definition}\label{def:decomp}
    (Decomposable Criterion). We say that a scoring criterion \(S(\mathbb{G}, D)\) is decomposable if it can be written as a sum of measures, each of which is a function only of one node and its parents. In other words, a decomposable scoring criterion \(\mathbb{S}\) applied to a DAG \(\mathbb{G}\) can always be expressed as:
\begin{equation}\label{eq:decomp}
    S(\mathbb{G}, \boldsymbol{D}) = \sum^n_i s(X_i, \textbf{Pa}^{\mathbb{G}}_i)
\end{equation}
\end{definition}

\begin{definition}[Score equivalent]\label{def:score_equivalent}\citet{ges}.
A score \(S\) is score equivalent if it assigns the same score to all the graphs in the same MEC.
\end{definition}
\begin{definition}[Local Consistency]\label{def:local_consistency}\citet{ges}
Let \(\boldsymbol{D}\) contain \(m\) iid samples from some distribution \(p(.)\). Let \(\mathbb{G}\) be any possible DAG and \(\mathbb{G}'\) a different DAG obtained by adding the edge \(i \rightarrow j\) to \(\mathbb{G}\). A score \(S\) is locally consistent if both hold: 
\begin{itemize}
    \item If \(X_i \not\perp_p X_j|Pa^{\mathbb{G}}_j, \) then \(S(\mathbb{G'}, \boldsymbol{D}) > S(\mathbb{G}, \boldsymbol{D})\)
    \item If \(X_i \perp_p X_j|Pa^{\mathbb{G}}_j, \) then \(S(\mathbb{G'}, \boldsymbol{D}) < S(\mathbb{G}, \boldsymbol{D})\)
\end{itemize}
\end{definition}

\subsection{Assumptions}
\begin{assumption}[Temporal Precedence]\label{assumption:temporal_precedence}
Given a perfectly recorded sequence of events \(((x_1, t_1), \cdots, (x_L, t_L))\) with labels \((\boldsymbol{y}_L, t_L)\) and monotonically increasing time of occurrence \(0 \leq t_1 \leq \cdots \leq t_L\), an event \(x_i\) is allowed to influence any subsequent event \(x_j\) such that \(t_i \leq t_j\) and \(i < j\). Formally, the graph \(\mathbb{G} = (\boldsymbol{U}, \boldsymbol{E})\), \( (x_i, x_j) \in \boldsymbol{E} \implies t_i \leq t_j \; \text{and step} \; i < j\)
\end{assumption}
It allows us to remove ambiguity in causal directionality and is widely used in time-series and sequential data \cite{cd_temporaldata_review}.

\begin{assumption}[Bounded Lagged Effects]\label{assumption:lagged_effects}
Once we observed events up to timestamp \(t_i\) and step \(i\) as \(\boldsymbol{Z}_{\leq t_i} = ((x_1, t_1), \cdots, (x_i, t_i))\), any future lagged copy of event \(X^{(t_i + \tau)}_i\) is independent of \(Y_j\) conditioned on \(\boldsymbol{Z}_{\leq t_i}\):
\[
Y_j \perp X^{(t_i + \tau)}_i | \boldsymbol{Z}_{\leq t_i}
\]
Where \(\tau = t_{i+1} - t_i\) is a finite bound on the allowed time delay for causal influence. 
\end{assumption}
In other words, we allow the causal influence of event \(X_i\) on \(Y_j\) until the next event \(X_{i+1}\) is observed. We note that for data with strong lagged effects (e.g., financial transactions), this might not hold well, but for log-based and error code-based data, this is usually correct.
\begin{assumption}[Causal Sufficiency for Labels]
\label{assumption:causal_sufficiency}
All relevant variables are observed, and there are no hidden confounders affecting the labels.
\end{assumption}

\begin{assumption}[Oracle Models]
\label{assumption:oracle} 
We assume that two autoregressive Transformer models, \(\text{Tf}_x\) and \(\text{Tf}_y\), are trained via maximum likelihood on a dataset of multi-labeled event sequences \(D = \{S^1_l, \cdots, S^m_l\} \subset \mathbb{S}\), and can perfectly approximate the true conditional distributions of events and labels:
\begin{equation}\label{eq:oracle}
    P(X_i|\text{Pa}(X_i)) = P_{\theta_x}(X_i|\text{Pa}(X_i)) = \text{Tf}_x(S_{< i}), \;
    P(Y_j|\text{Pa}(Y_j)) = P_{\theta_y}(Y_j|\text{Pa}(Y_j))=\text{Tf}_y(S_{\leq i})
\end{equation}
\end{assumption}
\subsection{Lemmas}

\begin{lemma}[Identifiability of \(\mathbb{G}\)]\label{lemma:oracle_identifiability}
    Assuming the faithfulness condition holds for the true causal graph \(\mathbb{G}\). Let \(\text{Tf}_x\) and \(\text{Tf}_y\) be oracle models that model the true conditional distributions of events and labels, respectively. The joint distribution \(P_{\theta_x, \theta_y}\) can then be constructed, and any conditional independence detected from the distributions estimated by \(\text{Tf}_x\) and \(\text{Tf}_y\) corresponds to a conditional independence in \(\mathbb{G}\):
    \[
    X_i \perp_{\theta_x, \theta_y} Y_j \mid \boldsymbol{Z} \quad \implies \quad X_i \perp_{\mathbb{G}} Y_j \mid \boldsymbol{Z}.
    \]
    Where \( \perp_{\theta_x, \theta_y}\) denotes the independence entailed by the joint probability \(P_{\theta_x, \theta_y}\).
\end{lemma}

\begin{lemma}[Markov Boundary Equivalence]\label{lemma:mb_par}
In a multi-label event sequence \(S_l\) and under the temporal precedence assumption A\ref{assumption:temporal_precedence}, the Markov Boundary of each label \(Y_j\) is only its parents such that \(\forall X \in \{\boldsymbol{U} - \text{Pa}(Y_j)\},  X \perp Y_j|\text{Pa}(Y_j) \Leftrightarrow \text{MB}(Y_j) = \text{Pa}(Y_j) \). 
\end{lemma}

\section{Proofs}\label{appendix:proofs}
We provide proofs for the results described in Section \ref{sec:oneshotmarkov}

\subsection{Proof of Lemma 1}\label{proof:lemma_one_faith}
\begin{proof}
We assume that the data is generated by the associated causal graph \(\mathbb{G}\) following the sequential BN from a multi-labelled sequence \(S\). And that the faithfulness assumption holds \cite{pearl_1998_bn}, meaning that all conditional independencies in the observational data are implied by the true causal graph \(\mathbb{G}\). 

Given that the Oracle models \(\text{Tf}_x\) and \(\text{Tf}_y\) are trained to perfectly approximate the true conditional distributions, for any variable \(U_i\) in the graph, we have:
\[
    P(U_i | \text{Pa}(U_i)) =
    \begin{cases}
        P(Y_j | \text{Pa}(Y_j)) = P_{\theta_y}(Y_j | \text{Pa}(Y_j)), & \text{if } U_i \in \boldsymbol{Y} \\
        P(X_i | \text{Pa}(X_i)) = P_{\theta_x}(X_i | \text{Pa}(X_i)), & \text{otherwise}.
    \end{cases}
\]
The joint distribution \(P_{\theta_x, \theta_y}\) can then be constructed using the chain rule \(P_{\theta_x, \theta_y}(X_1, \cdots, X_i, Y_1, \cdots, Y_c) = \prod^i_{k=0}P(X_k|Pa(X_k)) \prod^c_lP(Y_l|\text{Pa}(Y_l)\).
By the faithfulness assumption \cite{pearl_1998_bn}, if the conditional independencies hold in the data, they must also hold in the causal graph \(\mathbb{G}\): 
\[X_i \perp Y_j | \boldsymbol{Z} \implies X_i \perp_{\mathbb{G}} Y_j | \boldsymbol{Z}\]
Since we can approximate the true conditional distributions, it follows that:
\[X_i \perp_{\theta_x, \theta_y} Y_j |\boldsymbol{Z} \implies X_i \perp Y_j | \boldsymbol{Z} \implies X_i \perp_{\mathbb{G}} Y_j | \boldsymbol{Z}\]
Where \( \perp_{\theta_x, \theta_y}\) denotes the independence entailed by the joint probability \(P_{\theta_x, \theta_y}\).
Thus, the graph \(\mathbb{G}\) can be identified from the observational data.
\end{proof}
\subsection{Proof of Lemma 2}\label{proof:lemma_two_mb}
\begin{proof}
    Let \(<\boldsymbol{U}, \mathbb{G}, P>\) be the sequential BN composed of the events from the multi-labeled sequence \(S_l = (\{(t_1, x_1, \cdots, (t_L, x_L)\}^L_{i=1}, (\boldsymbol{y}_L, t_L))\). Following the 
temporal precedence assumption A\ref{assumption:temporal_precedence}, the labels \(\boldsymbol{y_L}\) can only be caused by past events \((x_1, \cdots, x_L)\); moreover, by definition, labels do not cause any other labels. Thus, \(Y_j\) has no descendants, so no children and spouses. Therefore, together with the Markov Assumption, we know that \(\forall X \in \{\boldsymbol{U} - Pa(Y_j)\}: Y_j \perp X|Pa(Y_j)\). Which is the definition of the MB (Def. \ref{def:markov_boundary}). Thus, \(\textbf{MB}(Y_j) = Pa(Y_j)\).

\end{proof}

\subsection{Proof of Theorem 1.}\label{proof:th1}
\begin{proof}
    By recurrence over the sequence length \(L\) of the multi-label sequence \(S^k_l\), we want to show that under temporal precedence A\ref{assumption:temporal_precedence}, bounded lagged effects A\ref{assumption:lagged_effects}, causal sufficiency A\ref{assumption:causal_sufficiency}, Oracle Models A\ref{assumption:oracle} the Markov Boundary of label \(Y_j\) can be identified in the causal graph \(\mathbb{G}\). 
    
Let's define \(\mathcal{M}^L_j\) as the estimated Markov Boundary of \(Y_j\) after observing \(L\) events.

\textbf{Base Case:} \(\mathbf{L=1}\):
Consider the BN for step \(L=1\) following the Markov assumption \cite{pearl_1998_bn} with two nodes \(X_1, Y_j\). Using \(\text{Tf}_x, \text{Tf}_y\) as Oracle Models A\ref{assumption:oracle}, we can express the conditional probabilities for any node \(U\):

\begin{equation}
    P(U | \text{Pa}(U)) =
    \begin{cases}
            P(X_1) = P_{\theta_x}(X_1 | [CLS]) \; \text{if}\; U \in \boldsymbol{X} \\
            P(Y_j|X_1) = P_{\theta_y}(Y_j | X_1)\; \text{otherwise}
    \end{cases}
\end{equation}

Assuming that P is faithful (A\ref{def:bn_faithfulness}) to \(\mathbb{G}\), no hidden confounders bias the estimate (A\ref{assumption:causal_sufficiency}) and temporal precedence (A\ref{assumption:temporal_precedence}), we can estimate the CMI \ref{eq:cmi_theorique} such that iif \(I(X_1, Y_j)|\emptyset) > 0 \Leftrightarrow Y_j \not\perp_{\theta_x, \theta_y} X_1 \implies Y_j \not\perp_{\mathbb{G}} X_1\) (Lemma \ref{lemma:oracle_identifiability}).

Since we assume temporal precedence A\ref{assumption:temporal_precedence}, we can orient the edge such that \(X_1\) must be a parent of \(Y_j\) in \(\mathbb{G}\). Using Lemma \ref{lemma:mb_par}, we know that \(Par(Y_j) = \textbf{MB}(Y_j) \implies X_{1} \in \textbf{MB}(Y_j)\), thus we must include \(X_1\) in \(M^1_j \), otherwise not. 

\textbf{Heredity:}
For \(L = i\), we obtained \(M^{i}_j\) with the sequential BN up to step \(L=i\). 
Now for \(L = i+1\), the sequential BN has \(i+2\) nodes denoted as \(\boldsymbol{U'} = (X_1, \cdots, X_i, X_{i+1}, Y_j)\). Using the Oracle Models A\ref{assumption:oracle} and following the Markov assumption \citep{pearl_1998_bn}, we can estimates the following conditional probabilities for any nodes \(U \in \boldsymbol{U'}\):

\begin{equation}
    P(U | \text{Pa}(U)) =
    \begin{cases}
        P(Y_j | \text{Pa}(Y_j)) \approx P_{\theta_y}(Y_j | \text{Pa}(Y_j)), & \text{if } U \in \boldsymbol{Y} \\
        P(X| \text{Pa}(X)) \approx P_{\theta_x}(X|\text{Pa}(X)), & \text{otherwise}.
    \end{cases}
\end{equation}
By bounded lagged effects (A\ref{assumption:lagged_effects}) we know that the causal influence of past \(X_{\leq i}\) on \(Y_j\) has expired. Moreover, no hidden confounders (A\ref{assumption:causal_sufficiency}) bias the independence testing.
Finally,
using Eq. \eqref{eq:cmi_theorique} we can estimate the CMI such that iif
\(I( Y_j, X_{i+1}| \boldsymbol{Z}) > 0 \Leftrightarrow Y_j \not\perp_{\theta_x, \theta_y}  X_{i+1} | \boldsymbol{Z} \implies Y_j \not\perp_{\mathbb{G}} X_{i+1} | \boldsymbol{Z}\) (Lemma \ref{lemma:oracle_identifiability}).

Since we assume temporal precedence A\ref{assumption:temporal_precedence}, we can orient the edge so that \(X_{i+1}\) must be a parent of \(Y_j\) in \(\mathbb{G}\). Using Lemma \ref{lemma:mb_par}, we know that \(Par(Y_j) = \textbf{MB}(Y_j) \implies X_{i+1} \in \textbf{MB}(Y_j)\).
Thus \(X_{i+1} \in M^{i+1}_j\) which represent the \textbf{MB}(\(Y_j)\) for step \(i+1\).

Finally, \(\mathcal{M}^{i+1}_j\) still recovers the Markov Boundary of \(Y_j\) such that \[\forall U \in \{\boldsymbol{U'} - \mathcal{M}^{i+1}_j\}, Y_j \perp U|\mathcal{M}^{i+1}_j\]
\end{proof}

\newpage
\section{Ablations}

\subsection{NADEs Quality.}\label{appendix:nades}
We did several ablations on the quality of the NADEs and their impact on the one-shot causal discovery phase. In particular, Table \ref{tab:ablation_nades_comparison} presents multiple \(\text{Tf}_x, \text{Tf}_y\) with respectively 90 and 15 million parameters or 34 and 4 million parameters. We also varied the context window (conditioning set \(\boldsymbol{Z}\)), trained on different amounts of data (tokens), and reported the classification results on the test set of \(\text{Tf}_y\) alone. We didn't output the running time since it was approximately the same for all NADEs: \(1.27\) minutes of 50,000 samples and \(0.14\) for 5000.

We observe that scaling up the NADEs model size and the trained data show the most significant improvements. Afterward, it is via the context \(c\), which, after \(c=15\), shows a decline in performance across a larger number of samples. We then choose the backbone with 1.5B Tokens, 105m parameters, and a context \(c=15\) for our experiments.
\begin{table}[ht]
\centering
\caption{Ablations of the performance of Phase 1 (One-shot \textbf{MB} retrieval) in function of different NADEs with \(m=50{,}000\) and \(m=500\) samples averaged over 6-folds. Classification metrics use weighted averaging. Metrics are given in \(\%\).}
\label{tab:ablation_nades_comparison}
\begin{tabular}{lcccccc}
\toprule
\textbf{Tokens} & \textbf{Parameters} & \textbf{Context} & \textbf{Precision (↑)} & \textbf{Recall (↑)} & \textbf{F1 Score (↑)} & \textbf{Tfy F1 (↑)} \\
\midrule
\multicolumn{7}{c}{\textit{For \(n = 50{,}000\) samples}} \\
\midrule
1.5B & 105m & \(c = 4\)  & \(47.95 \pm 1.05\) & \(30.65 \pm 0.51\) & \(37.39 \pm 0.67\) & 88.6 \\
1.5B & 105m & \(c = 12\) & \(54.62 \pm 1.03\) & \(29.88 \pm 0.73\) & \(38.63 \pm 0.85\) & 90.43 \\
1.5B & 105m & \(c = 15\) & \(\mathbf{55.26 \pm 1.42}\) & \(\mathbf{31.37 \pm 0.82}\) & \(\mathbf{40.02 \pm 1.03}\) & 90.57 \\
1.5B & 105m & \(c = 20\) & \(49.52 \pm 1.59\) & \(\mathbf{31.76 \pm 0.85}\) & \(36.54 \pm 1.10\) & 91.19 \\
1.5B & 105m & \(c = 30\) & \(36.65\pm 1.18\) & \(22.75 \pm 0.78\) & \(26.57 \pm 0.91\) & \textbf{92.64} \\
300m & 47m & \(c = 20\) & \(39.49 \pm 1.77\) & \(26.30 \pm 0.89\) & \(29.01 \pm 1.10\) & 83.6 \\
\midrule
\multicolumn{7}{c}{\textit{For \(n = 500\) samples}} \\
\midrule
1.5B & 105m & \(c = 12\) & \(54.84 \pm 4.55\) & \(\mathbf{31.45 \pm 2.23}\) & \(\mathbf{39.95 \pm 2.83}\) & 90.43 \\
1.5B & 105m & \(c = 15\) & \(55.04 \pm 3.36\) & \(29.90 \pm 1.78\) & \(38.74 \pm 2.24\) & 90.57 \\
1.5B & 105m & \(c = 20\) & \(48.84 \pm 4.01\) & \(\mathbf{31.65 \pm 2.37}\) & \(36.19 \pm 2.65\) & \textbf{91.19} \\
300m & 47m & \(c = 20\) & \(38.23 \pm 2.91\) & \(25.31 \pm 2.39\) & \(27.92 \pm 2.25\) & 83.6 \\
\bottomrule
\end{tabular}
\end{table}

\subsection{Sampling Number}\label{sec:samplingnumber}
\begin{figure}[!h]
    \centering
    \caption{Evolution of several classification metrics (one-shot) and elapsed time per sample in function of the number of samples \(N\) chosen. Results are reported using 1-sigma error bar.}
    \includegraphics[width=0.8\linewidth]{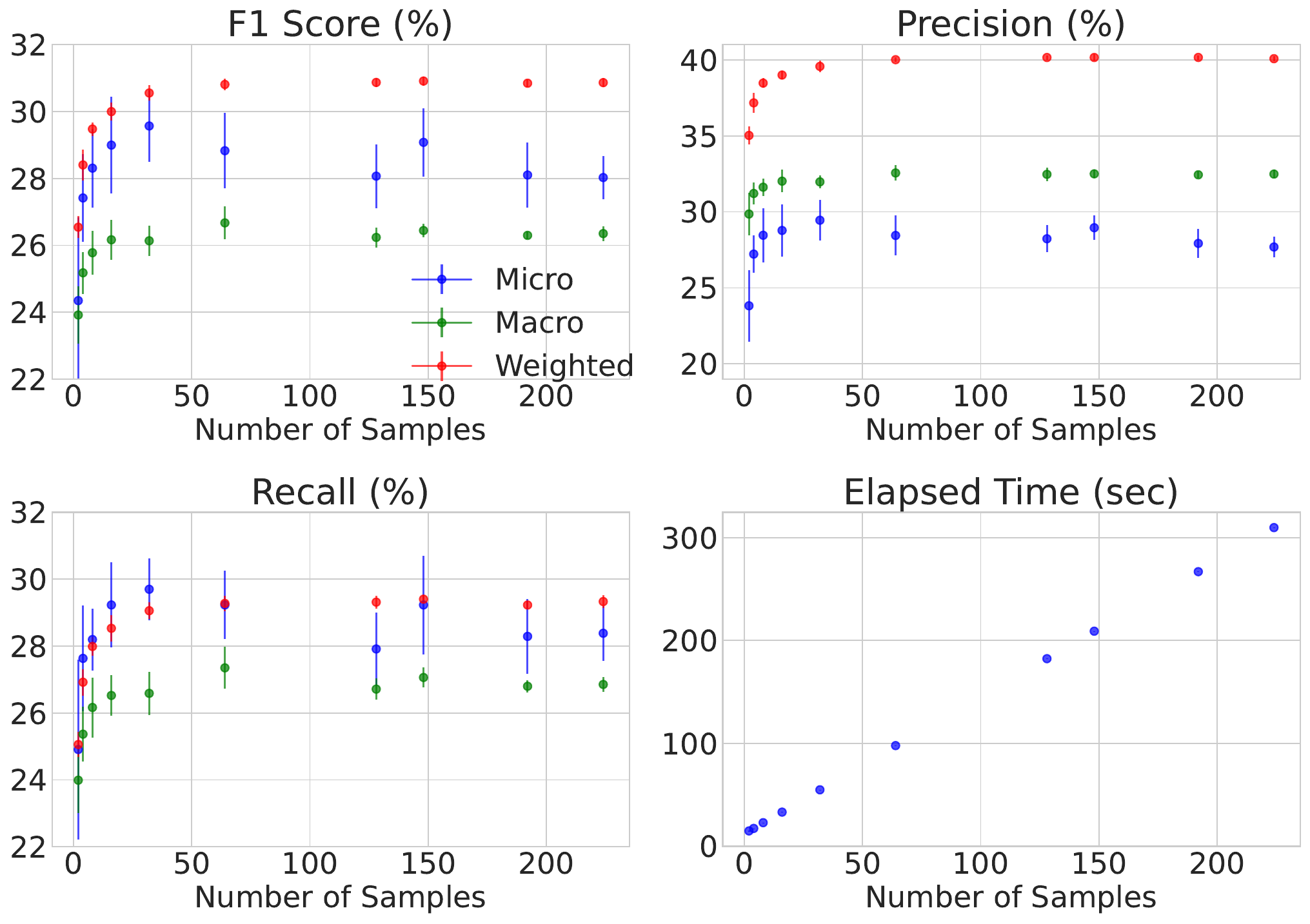}
    \label{abl:number_sampling}
\end{figure}

We tested different values of \(N\) for the sampling method across averaging methods (micro, macro, weighted), as shown in Fig. \ref{abl:number_sampling}. We performed eight different runs and reported the average, standard deviation, and elapsed time. In general, sampling with a larger \(N\) tends to reduce the standard deviation and yield more reliable Markov Boundary estimates. Moreover, as we process more samples, the model gradually improves, following a logarithmic growth pattern until it converges to a final score. We also verify that our time complexity is linear with the number of samples \(N\). 
Based on these results, we generally select \(N=68\) as the sample size.

\subsection{Dynamic Thresholding}\label{sec:dynamic_threshold}
\begin{figure}[!ht]
    \centering
    \caption{Evolution of one-shot F1 Score, Precision, and Recall in function of coefficient \(k\). Results  are reported using 1-sigma error bar.}
    \label{abl:threshold_selection}
    \includegraphics[width=0.8\linewidth]{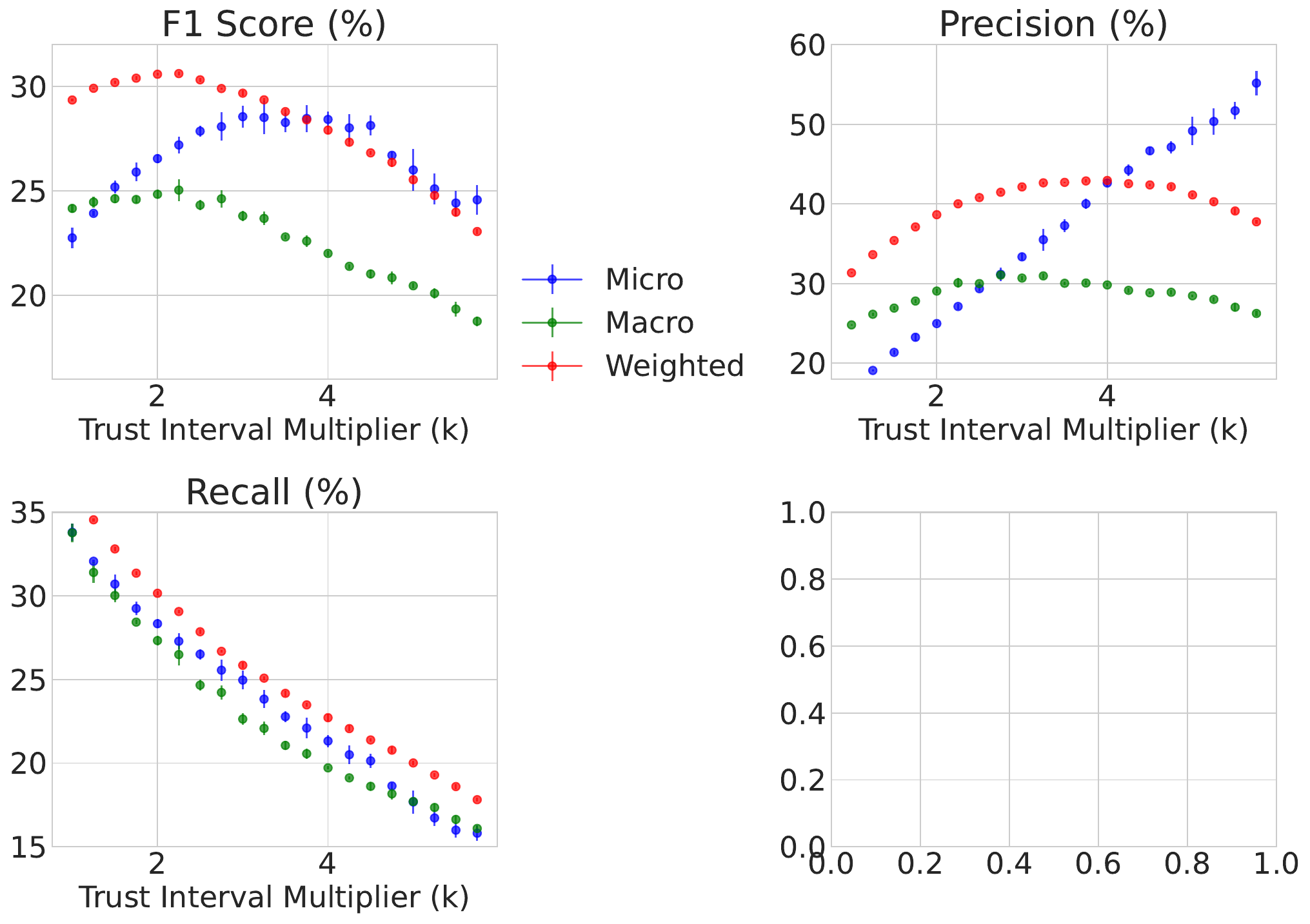}
\end{figure}

We performed ablations on the effect of \(k\) during the dynamic thresholding of the CMI (Eq. \eqref{eq:cmi_epsilon}) to access conditional independence in Fig. \ref{abl:threshold_selection}. To balance the classification metrics across the different averaging, we set \(k=2.75\).

\newpage
\subsection{Criterions}\label{appendix:criterions}
This section presents the different criteria used for comparison in the experimental evaluation.

\subsection{Frequency}
Frequency-based heuristics that apply fixed thresholds \(\tau\) to the empirical frequency of the occurrence of \(X_i\) in each of the Markov Boundary \(\textbf{MB}(Y_j)\). Formally, 
For each label \(Y_j\), after merging all local edge sets into a global set \(E = \bigcup^m_{i=1} E_i\), we evaluate each candidate variable \(X_i \in \textbf{MB}_j\) based on its frequency of appearance across the local models. If this frequency exceeds the threshold \(\tau\), the variable is retained in the final merged \(\textbf{MB}_j\); otherwise, it is discarded.

\subsubsection{Expected FPR Adjustment}
Same principle as in \cite{consensus_bn_freq_cutoff_no} except that we fit the two Beta distributions on the mutual information of \(I(Y_j, X_i)\) instead of the raw frequencies.

\subsubsection{Mutual Information}

\paragraph{BES}
The BES is the second phase of GES \cite{ges}, where edges are removed one after the other to maximize a criterion \(S\).
Heuristics approaches \cite{approx_fusionPUERTA2021155, delSagrado2001} aim to solve this problem by optimizing: 
\begin{equation}\label{eq:criterion_edges}
E = \underset {E_i \in \varepsilon}\arg {\text{max}} \sum_{e \in E_i} S(e)    
\end{equation}
Where \(\varepsilon\) denotes the search space (all possible edges over \(\boldsymbol{U}\)) and \(S(e)\) a criterion function for edge relevance (e.g. edge frequency, thresholds, \(\cdots\)). 
This formulation takes into account the underlying edges' characteristics but not the overall network structure, complexity and missing data \cite{consensus_bn_jose}, leading to a \textit{consensus fusion approach} \cite{torrijos2025informedgreedyalgorithmscalable}.

\paragraph{Estimating Mutual Information in Event Sequences.}
We want to reuse the estimated conditional mutual information, Eq. \eqref{eq:cmi_theorique}, and profit from the parallelized inference of Phase 1 (Fig. \ref{fig:cargo}).

As argued out by \citet{JanzingDominik2013QCI}, a causal strength measure (or criterion) \(C_{X_i \rightarrow Y_j}\) should possess multiple properties. Notably, if \(C_{X_i \rightarrow Y_j} = 0\),  then the joint distribution satisfies the Markov condition with respect to the DAG obtained by removing
the arrow \({X_i \rightarrow Y_j}\). Moreover, the true DAG reads \(X_i \rightarrow Y\) iff \(C_{X_i \rightarrow Y_j} = I(X_i, Y_j)\). 

It is a natural criterion for merging edges across multiple causal graphs. However, it remains tricky to estimate \cite{estimating_mi}.
Using the chain rule of conditional mutual information \cite{cover1999elements}, we can rewrite it as:
\begin{equation}\label{eq:mi}
    I(Y_j, X_i|\boldsymbol{Z}) = I(Y_j, X_i) - I(Y_j, X_i, \boldsymbol{Z})
\end{equation}
Where  \(I(Y_j, X_i, \boldsymbol{Z})\) is the interaction information \cite{cover2006elements}, which tells us whether knowing \(\boldsymbol{Z}\) explains away the dependency between \(X_i\) and \(Y_j\) (negative interaction), or enhances it (positive interaction):

\[ I(Y_j, X_i, \boldsymbol{Z}) \triangleq I(Y_j, \boldsymbol{Z}) - I(Y_j, \boldsymbol{Z}|X_i)\]

\(I(Y_j, \boldsymbol{Z})\) can be estimated using the same Monte-Carlo sampling as for \(I(Y_j, X_i|\boldsymbol{Z})\) \eqref{eq:cmi_theorique}. 
Since \(I(Y_j, \boldsymbol{Z})=H(Y_j) - H(Y|\boldsymbol{Z_j})\), the marginal \(p(y)\) is needed.
Fortunately, the dataset \(\boldsymbol{D}\) is large enough, hence the frequencies of \(y_j\) are recovered empirically and an estimate \(\hat{p}(y)\) which we assume to be equal to the true marginal \(p(y)\).
We acknowledge that under a restricted dataset, \(\hat{p}(y)\) might differ from \(p(y)\). This yields to: 
\begin{equation}\label{eq:mi_y_z}
    I(Y_j, \boldsymbol{Z}) = \mathbb{E}_{z} D_{KL}(P(Y_j|\boldsymbol{Z})||\hat{P}(Y_j)) = \mathbb{E}_{z} I_G(Y_j, z)
\end{equation}
Formally, we assume that for long sequences i.e  \(i \rightarrow +\infty\), our event sequences form a stationary ergodic stochastic process and \(I(Y_j, \boldsymbol{Z}| X_i)\) is negligible compare to \( I(Y_j, \boldsymbol{Z})\) since \(\boldsymbol{Z}\) is containing most of the information to predict \(Y_j\). This reduces the mutual information to 
\[I(Y_j, X_i) \approx I(Y_j, X_i|\boldsymbol{Z}) + I(Y_j|\boldsymbol{Z})\]

\paragraph{Criterion.}
We propose a \textbf{Class-Aware Information Gain (CAIG)} score for evaluating candidate edges during Phase 2 of CARGO. Given \(m\) i.i.d. samples from a dataset \(\boldsymbol{D}\), CAIG balances three key factors: mutual information derived from information gain, class imbalance, and network complexity. 

For each label node \(Y_j\), with candidate parent set \(\textbf{Pa}^{\mathbb{G}}_j\), the CAIG score is:
\begin{equation}\label{eq:caig}
    S(\mathbb{G}', \boldsymbol{D}) = \sum_{j=1}^{n} s_I(Y_j, \textbf{Pa}^{\mathbb{G}}_j) - \alpha \cdot |\textbf{Pa}^{\mathbb{G}} _j|\cdot \log{\left(\frac{m}{m_j} + 1\right)}
\end{equation}    
\text{With} \(s_I(Y_j, \textbf{Pa}^{\mathbb{G}}_j) = \sum_{X_i \in \textbf{Pa}^{\mathbb{G}}_j} I(Y_j, X_i)\),
\(\alpha\) is a regularization hyperparameter, \(m_j\) is the number of positive instances for class \(Y_j\).

This formulation encourages informative yet parsimonious graph structures, correcting for underrepresented labels via the regularization term. It is also efficient since CAIG is \emph{decomposable} \cite{ges} like BIC with the local \(s_I\). This criterion is denoted as \emph{BES mi imbalance par} in our experiments in Fig. \ref{fig:ablation_comparaison_criterions}.

\section{Implementation}\label{appendix:code}

\subsection{Computation.}
A key advantage of our approach is its scalability. Unlike traditional methods whose complexity depends on the event and label cardinality \(|\mathbb{X}|\) and \(|\mathbb{Y}|\) \cite{feature_selection_review}, our method is agnostic to both. 
As illustrated in Figure~\ref{fig:cargo}, all steps are parallelized on GPUs. CMI estimations are independently performed for all positions \(i \in [c, L]\), with the sampling pushed into the batch dimension and results averaged across labels, leading to $\text{BS} \times N \times L$ CI-tests per batch \(D = \{S^0_l, \ldots, S^m_l\}\). Consequently, time complexity transitions from \(\mathcal{O}(\text{BS} \times N \times L)\) to \(\mathcal{O}(1)\) per batch due to GPU parallelism. The complexity is bounded by the Transformers' inference part, where it scales quadratically with the sequence length  \(\mathcal{O}(L^2)\) if one uses vanilla self-attention \cite{tf}. 

\subsection{Phase 1}\label{sec:phase1implementation}
The following is the implementation of the one-shot phase in PyTorch \cite{pytorch}.
\begin{lstlisting}[language=Python, label={lst:phase1}]
def topk_p_sampling(z, prob_x, c: int, n: int = 64, p: float = 0.8, k: int = 35,
                       cls_token_id: int = 1, temp: float = None):
    # Sample just the context
    input_ = prob_x[:, :c]

    # Top-k first
    topk_values, topk_indices = torch.topk(input_, k=k, dim=-1)

    # Top-p over top-k values
    sorted_probs, sorted_idx = torch.sort(topk_values, descending=True, dim=-1)
    cum_probs = torch.cumsum(sorted_probs, dim=-1)
    mask = cum_probs > p
    
    # Ensure at least one token is kept
    mask[..., 0] = 0

    # Mask and normalize
    filtered_probs = sorted_probs.masked_fill(mask, 0.0)
    filtered_probs += 1e-8  # for numerical stability
    filtered_probs /= filtered_probs.sum(dim=-1, keepdim=True)

    # Unscramble to match the original top-k indices
    # Need to reorder the sorted indices back to the original top-k
    reorder_idx = torch.argsort(sorted_idx, dim=-1)
    filtered_probs = torch.gather(filtered_probs, -1, reorder_idx)

    batched_probs = filtered_probs.unsqueeze(1).repeat(1, n, 1, 1)        # (bs, n, seq_len, k)
    batched_indices = topk_indices.unsqueeze(1).repeat(1, n, 1, 1)        # (bs, n, seq_len, k)

    sampled_idx = torch.multinomial(batched_probs.view(-1, k), 1)         # (bs*n*seq_len, 1)
    sampled_idx = sampled_idx.view(-1, n, c).unsqueeze(-1)

    sampled_tokens = torch.gather(batched_indices, -1, sampled_idx).squeeze(-1)
    sampled_tokens[..., 0] = cls_token_id

    # Reconstruct full sequence
    z_expanded = z.unsqueeze(1).repeat(1, n, 1)[..., c:]
    return torch.cat((sampled_tokens, z_expanded), dim=-1)

from torch import nn
def OneShotCD(tfe: nn.Module, tfy: nn.Module, batch: dict[str, torch.Tensor], c: int, n: int, eps: float=1e-6, topk: int=20, k: int=2.75, p=0.8) -> torch.Tensor:
    """ tfe, tfy: are the two autoregressive transformers (event type and label)
        batch: dictionary containing a batch of input_ids and attention_mask of shape (bs, L) to explain.
        c: scalar number defining the minimum context to start inferring, also the sampling interval.
        n: scalar number representing the number of samples for the sampling method.
        eps: float for numerical stability
        topk: The number of top-k most probable tokens to keep for sampling
        k: Number of standard deviations to add to the mean for dynamic threshold calculation
        p: Probability mass for top-p nucleus
    """
    o = tfe(attention_mask=batch['attention_mask'], input_ids=batch['input_ids'])['prediction_logits'] # Infer the next event type
    x_hat = torch.nn.functional.softmax(o, dim=-1)

    b_sampled = topk_p_sampling(batch['input_ids'], x_hat, c, k=topk, n=n, p=p) # Sampling up to (bs, n, L)
    n_att_mask = batch['attention_mask'].unsqueeze(1).repeat(1, n, 1)

    with torch.inference_mode():
        o = tfy(attention_mask=n_att_mask.reshape(-1, b_sampled.size(-1)), input_ids=b_sampled.reshape(-1, b_sampled.size(-1))) # flatten and infer
        prob_y_sampled = o['ep_prediction'].reshape(b_sampled.size(0), n, batch['input_ids'].size(-1)-c, -1) # reshape to (bs, n, L-c)

        # Ensure probs are within (eps, 1-eps)
        prob_y_sampled = torch.clamp(prob_y_sampled, eps, 1 - eps)

        y_hat_i = prob_y_sampled[..., :-1, :] # P(Yj|z)
        y_hat_iplus1 = prob_y_sampled[..., 1:, :] # P(Yj|z, x_i) 

        # Compute the CMI & CS and average across sampling dim
        cmi = torch.mean(y_hat_iplus1*torch.log(y_hat_iplus1/y_hat_i)+ (1-y_hat_iplus1)*torch.log((1-y_hat_iplus1)/(1-y_hat_i)), dim=1)
        # (BS, L, Y)
        cs = y_hat_iplus1 - y_hat_i
        cs_mean = torch.mean(cs, dim=1)
        cs_std = torch.std(cs, dim=1)

        # Confidence interval for threshold
        mu = cmi.mean(dim=1)
        std = cmi.std(dim=1)
        dynamic_thresholds = mu + std * k

        # Broadcast to select an individual dynamic threshold
        cmi_mask = cmi >= dynamic_thresholds.unsqueeze(1)

        cause_token_indices = cmi_mask.nonzero(as_tuple=False)
        # (num_causes, 3) --> each row is [batch_idx, position_idx, label_idx]
        return cause_token_indices, cs_mean, cs_std, cmi_mask
\end{lstlisting}

\begin{remark}
    Since \textit{tfy} contains tfe as backbone, in practice we need only one forward pass from tfy and extract also \(\hat{\boldsymbol{x}}\), so \textit{tfe} is not needed. We let it to improve understanding and clarity.
\end{remark}
\subsection{Phase 2.}\label{sec:phase2implementation}
\begin{lstlisting}[language=Python, label={lst:phase2}]
import random
def create_auto_adaptive_threshold_fn(all_m_j, tau_max=0.5, tau_min=0.05, k=None, m0="median"):
    m_0 = np.median(all_m_j)

    if k == None:
        q25, q75 = np.percentile(all_m_j, [25, 75])
        if q75 == q25:
            k = 1.0
        else:
            log_iqr = np.log(q75) - np.log(q25)
            k = (2 * np.log(3)) / log_iqr

    def threshold_function(m_j):
        log_m_j = np.log(m_j + 1e-9)
        log_m_0 = np.log(m_0)
        logistic_decay = 1 / (1 + np.exp(k * (log_m_j - log_m_0)))
        return (tau_max - tau_min) * logistic_decay + tau_min

    return threshold_function

def adaptive_thresholding_frequency(graphs: list,
                  present_labels: dict,
                  frequency_threshold: float = 0.5,
                  k: float=None,
                  tau_min: float=0.05,
                  tau_max: float=0.5,
                  m0: str="median",
                  verbose=False,
                  **kwargs):
    """
    Frequency voting: keep edges appearing with frequency > threshold across samples.
    
    :param graphs: list of local graphs (e.g., from Phase 1). Each graph is a dict[label][token] = list of stats.
    :param present_labels: labels present in evaluation
    :param frequency_threshold: e.g. 0.5 for majority, 0.8 for conservative
    :return: filtered_labels, sample_per_label, elapsed_time
    """
    start_time = datetime.now()
    # Step 1: Aggregate graphs
    labels, sample_per_label = union(graphs)  # user-defined union function
    old_labels = labels.copy()
    nodes = count_nodes(labels)
    samples = len(graphs)

    # Create threshold function
    auto_threshold_fn = create_auto_adaptive_threshold_fn(list(sample_per_label.values()), k=k, tau_max=tau_max, tau_min=tau_min, m0=m0)

    # Step 2: Frequency voting with dynamic thresholds
    edge_counts = defaultdict(lambda: defaultdict(int))  # edge_counts[label][token] = count
    
    for g in graphs:
        for label, token_dict in g.items():
            if label not in labels:
                continue
            for token in token_dict:
                edge_counts[label][token] += 1

    # Step 3: Keep edges above frequency threshold
    filtered_labels = defaultdict(dict)
    for label in labels:
        total = sample_per_label.get(label, samples)  # fallback to total graphs if missing
        for token, count in edge_counts[label].items():
            freq = count / total
            if freq >= auto_threshold_fn(sample_per_label.get(label, 1)):
                filtered_labels[label][token] = {'frequency': freq}
                if verbose:
                    print(f"[{label}] token {token} kept (freq={freq:.2f})")

    nb_of_edges = sum(len(v) for v in filtered_labels.values())
    print(f"Time: {(datetime.now() - start_time).total_seconds():.2f}s")
    return filtered_labels, sample_per_label, (datetime.now() - start_time).total_seconds()

\end{lstlisting}

\section{Figures}

\begin{figure}[!h]
    \centering
    \includegraphics[width=1\linewidth]{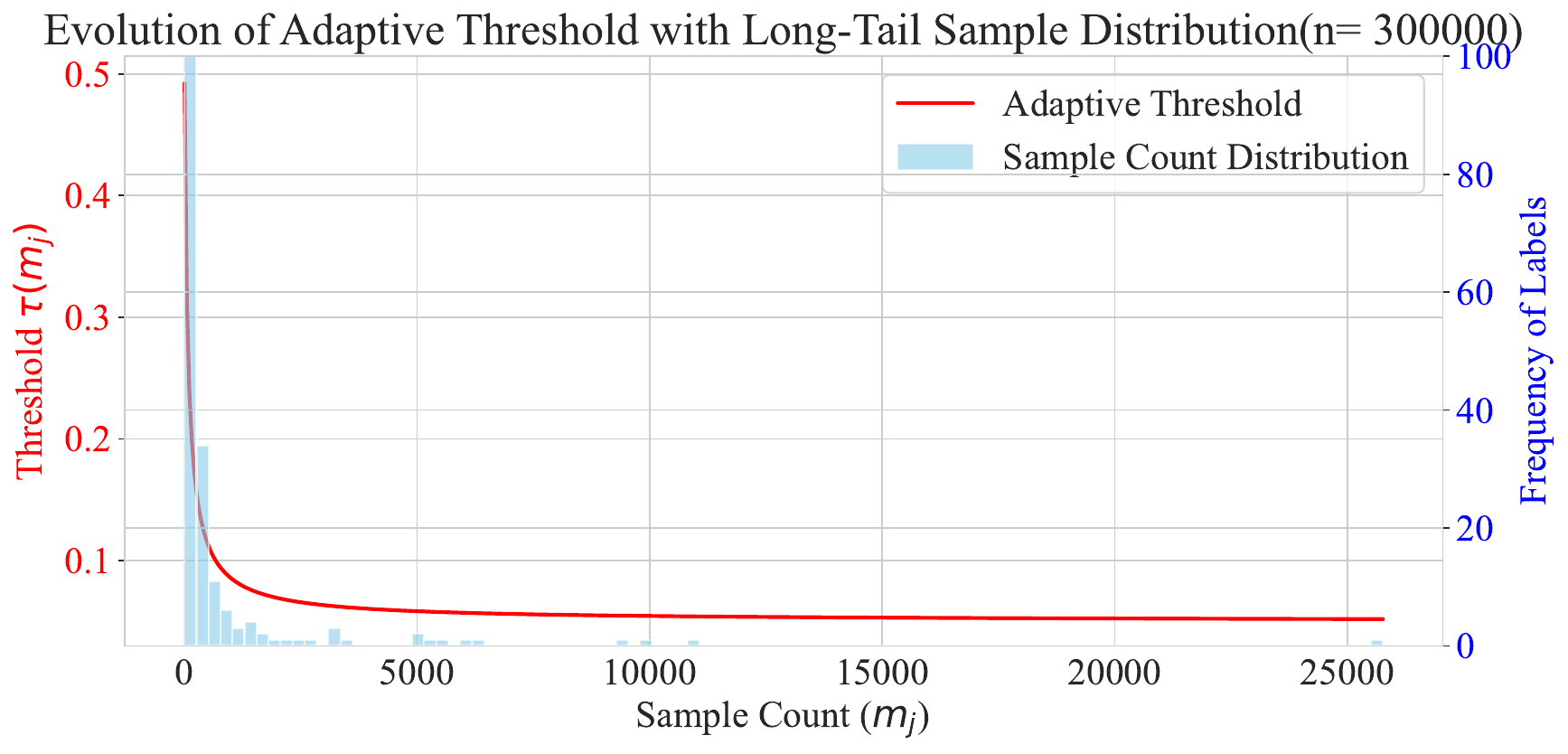}
    \caption{Adaptive thresholding function \(\tau_j(m_j)\) across varying label frequencies \(m_j\), illustrating the logistic decay from \(\tau_{\max}\) to \(\tau_{\min}\).}
    \label{fig:adaptive_threshold}
\end{figure}

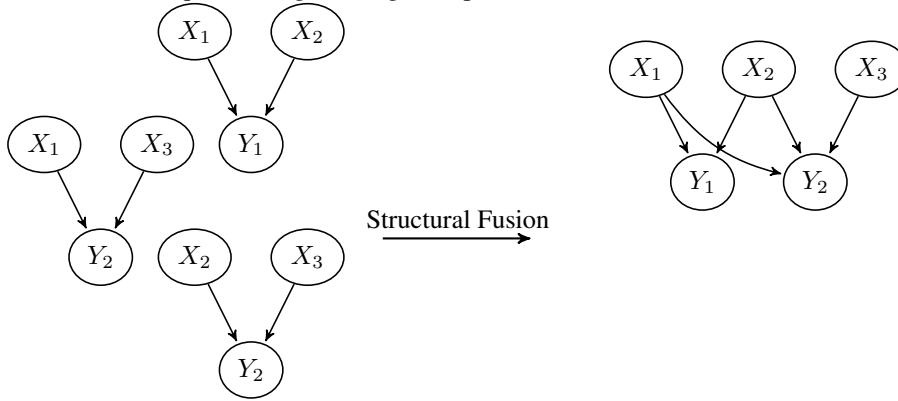
\begin{figure}[!h]
    \centering
    \caption{Illustration of structural fusion: individual causal graphs (left) aggregated into a fused DAG for multi-label event sequences (right) using a simple union.}
    \label{fig:structural_fusion_example}
    \begin{tikzpicture}[->,>=stealth',shorten >=1pt,auto,node distance=1.5cm,
                        semithick, every state/.style={circle,draw,minimum size=1.2em}]
        
        \node[state] (a1) at (0,2) {\(X_1\)};
        \node[state] (b1) at (1.5,2) {\(X_2\)};
        \node[state] (c1) at (0.75,0.5) {\(Y_1\)};
        \path (a1) edge (c1);
        \path (b1) edge (c1);

        \node[state] (a2) at (0,-1) {\(X_2\)};
        \node[state] (b2) at (1.5,-1) {\(X_3\)};
        \node[state] (c2) at (0.75,-2.5) {\(Y_2\)};
        \path (a2) edge (c2);
        \path (b2) edge (c2);

        \node[state] (a3) at (-2,0.5) {\(X_1\)};
        \node[state] (b3) at (-0.5,0.5) {\(X_3\)};
        \node[state] (c3) at (-1.25,-1) {\(Y_2\)};
        \path (a3) edge (c3);
        \path (b3) edge (c3);

        \draw[thick,->] (2.5,-0.75) -- (4.5,-0.75) node[midway,above] {Structural Fusion};

        \node[state] (x1) at (6,1.5) {\(X_1\)};
        \node[state] (x2) at (7.5,1.5) {\(X_2\)};
        \node[state] (x3) at (9,1.5) {\(X_3\)};
        \node[state] (y1) at (6.75,0) {\(Y_1\)};
        \node[state] (y2) at (8.25,0) {\(Y_2\)};

        \path (x1) edge (y1);
        \path (x2) edge (y1);
        \path (x2) edge (y2);
        \path (x3) edge (y2);
        \path (x1) edge[bend right=20] (y2);

    \end{tikzpicture}
\end{figure}
\end{document}